\documentclass[letterpaper, 10 pt, conference]{ieeeconf}  
\pdfoutput=1

\IEEEoverridecommandlockouts                              

\usepackage{subfigure}
\usepackage{siunitx}
\usepackage{epsfig} 
\usepackage[linesnumbered,ruled]{algorithm2e}

\newcommand{\eg}{{\em e.g.,~}}
\newcommand{\ie}{{\em i.e.,~}}
\newcommand{\etal}{\textit{et al.}~}

\title{\LARGE \bf
Stroke-based Rendering and Planning for Robotic Performance of Artistic Drawing
}

\author{Ivaylo Ilinkin$^{1}$, Daeun Song$^{2}$ and Young J. Kim$^{2}$
\thanks{$^{1}$I. Ilinkin is with the department of computer science at Gettysburg College in the U.S.A.  {\tt\small iilinkin@gettysburg.edu}}%
\thanks{$^{2}$D. Song and Y. J. Kim are with the department of computer science and engineering at Ewha womans university in Korea  {\tt\small daeunsong@ewhain.net, kimy@ewha.ac.kr}}%
}

\begin{document}

\maketitle
\thispagestyle{empty}
\pagestyle{empty}

\begin{abstract}



We present a new robotic drawing system based on stroke-based rendering (SBR). Our motivation is the artistic quality of the whole performance. Not only should the generated strokes in the final drawing resemble the input image, but the stroke sequence should also exhibit a human artist's planning process. Thus, when a robot executes the drawing task, both the drawing results and the way the robot executes would look artistic. 

Our SBR system is based on image segmentation and depth estimation. It generates the drawing strokes in an order that allows for the intended shape to be perceived quickly and for its detailed features to be filled in and emerge gradually when observed by the human. This ordering represents a stroke plan that the drawing robot should follow to create an artistic rendering of images. We experimentally demonstrate that our SBR-based drawing makes visually pleasing artistic images, and our robotic system can replicate the result with proper sequences of stroke drawing.

\end{abstract}

\section{INTRODUCTION}\label{sec:intro}

The early incarnation of mechanical drawing can be traced back several decades to Tinguely and AARON \cite{aaron90}. Since then, many exciting approaches for robotic drawing, or more broadly robotic art, have been pursued by roboticists. Thanks to the availability of robust and affordable robotic and sensing hardware components, more and more modern artists and scientists have been engaged in pushing the frontier of robotic art. This research direction has been fueled further by recent progress in generative art in machine learning and fast and scalable computer graphic algorithms.

Most of the existing robotic art systems focus on mimicking the drawings of human artists using robots. The conventional robotics paradigm of sensing, planning, and control has been exploited to autonomously create these drawings as accurately or quickly as possible. Consequently, robotic drawing has traditionally been evaluated based on how close the drawing result is to human art. In this paper, we divert this research trend by asking {\em how} the human artist would draw rather than {\em what} the human artist would draw. In other words, we would like to create an autonomous robotic drawing system that not only creates an artistic drawing but also executes the task in a manner that resembles how a human artist would draw. This question boils down to the planning aspect of the robotics paradigm.

\begin{figure}[htb]
\centering
\includegraphics[height=3.4cm]{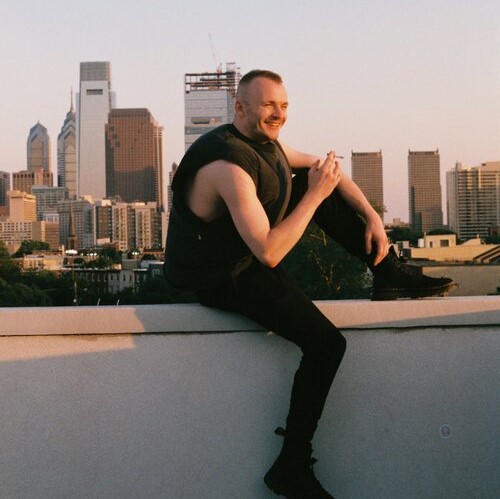}
\includegraphics[height=3.4cm]{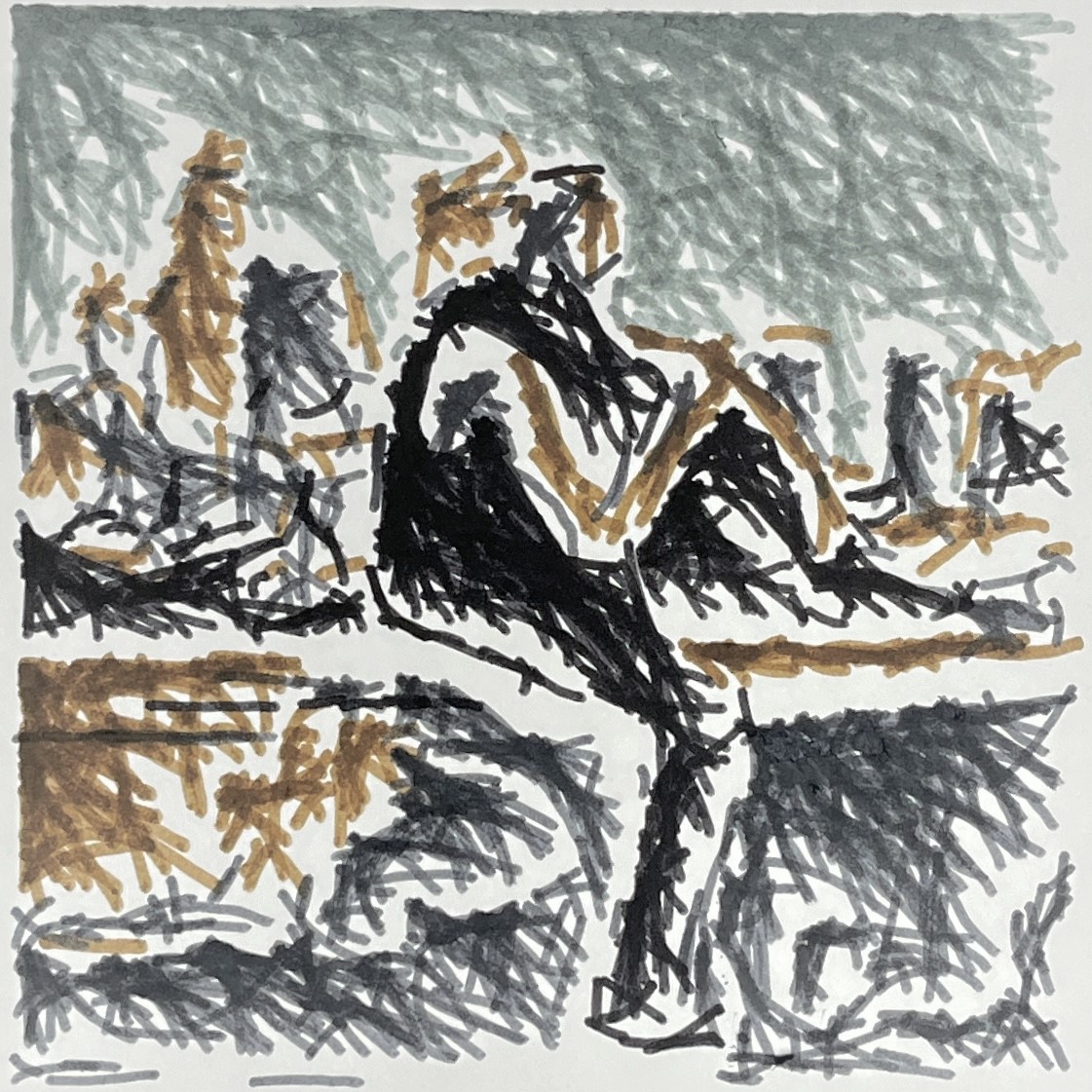}
\includegraphics[height=3.4cm]{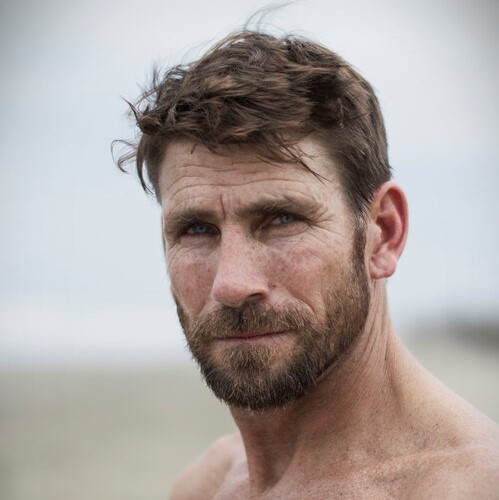}
\includegraphics[height=3.4cm]{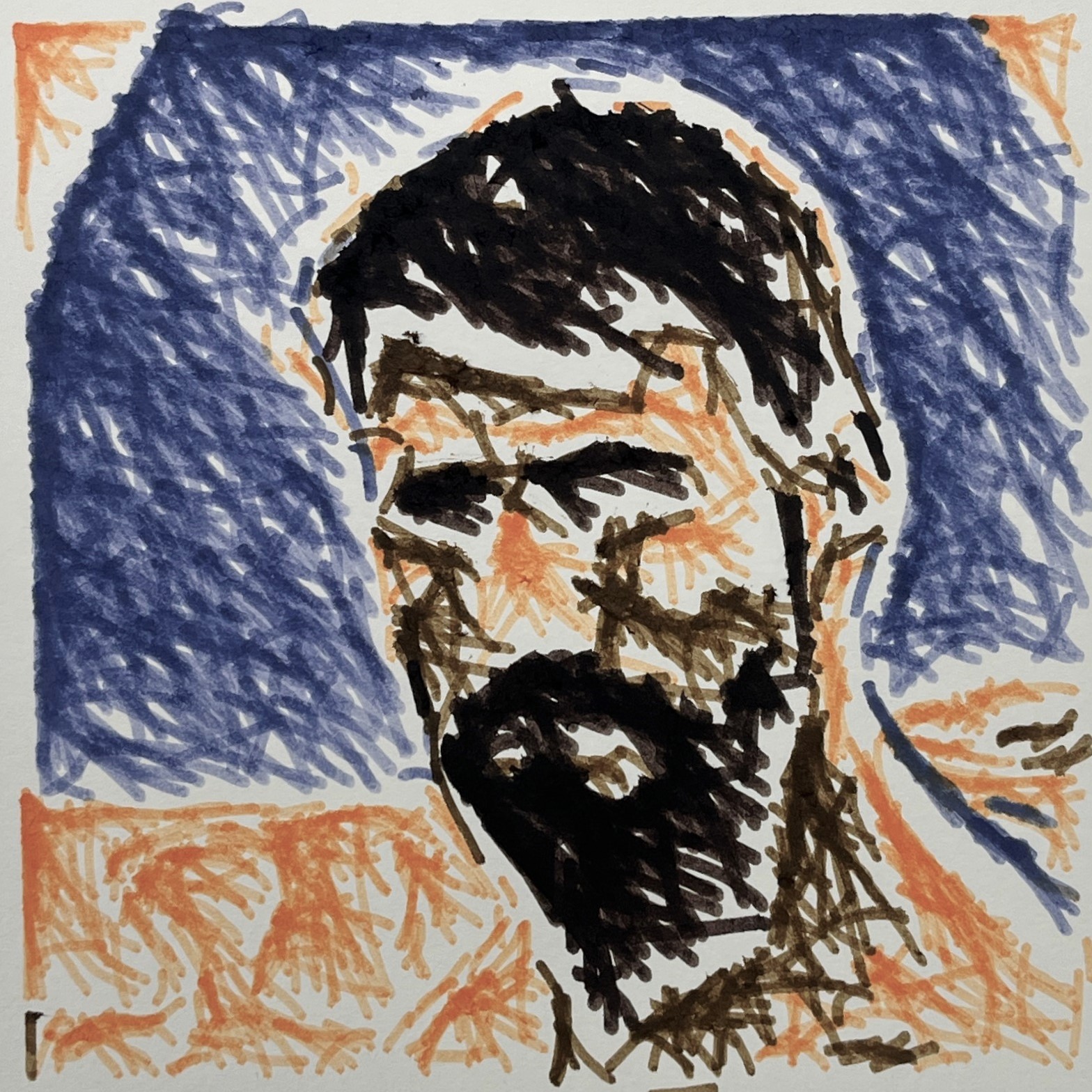}
\includegraphics[height=2.74cm]{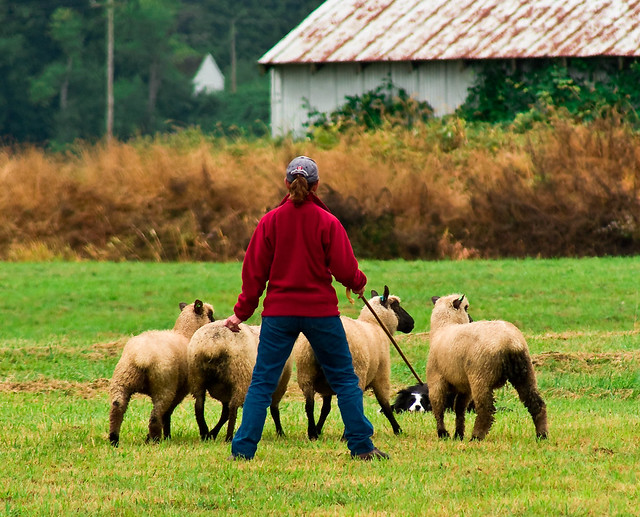}
\includegraphics[height=2.74cm]{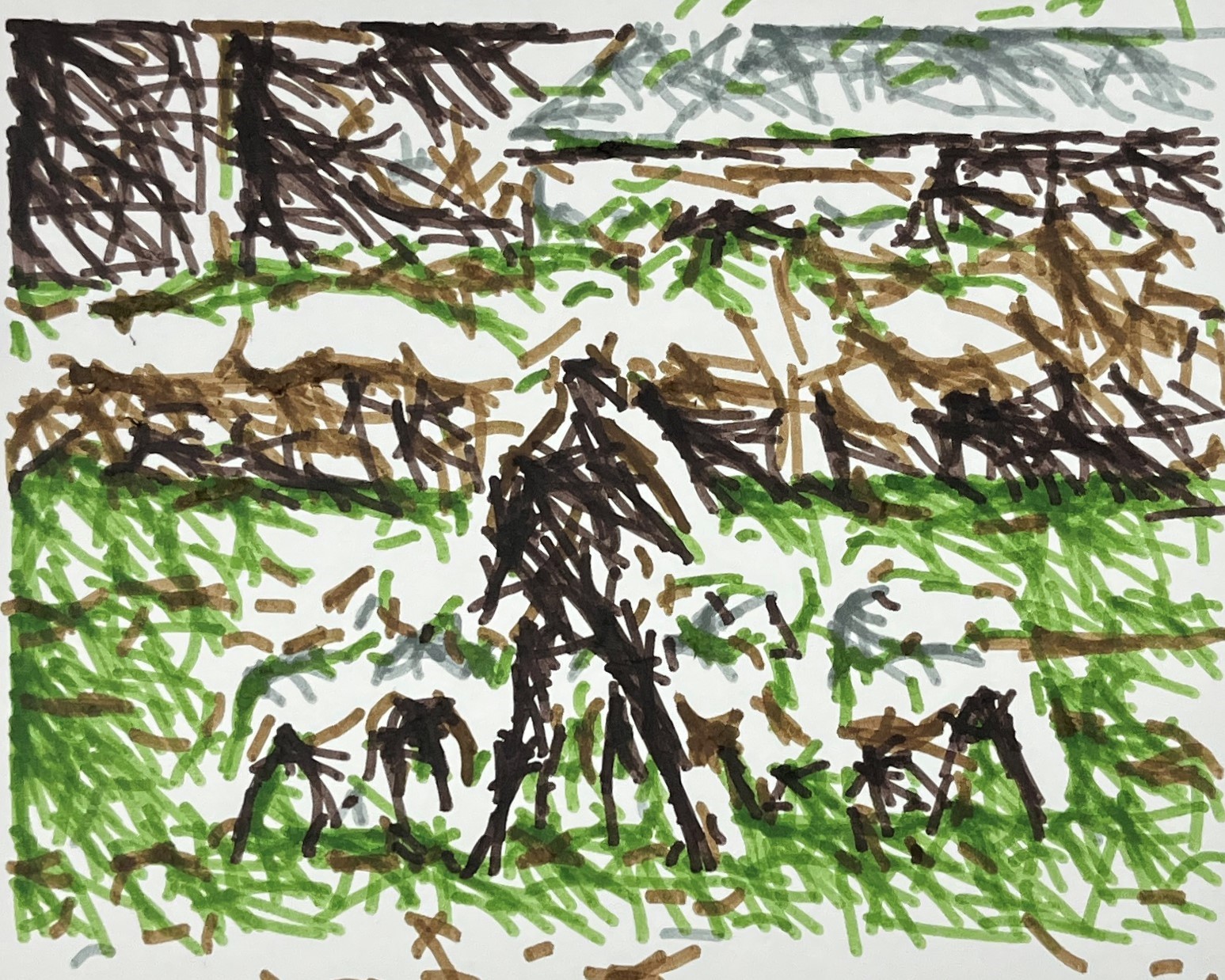}
\caption{Stroke-based robotic drawings. Left: original images. Right: our robotic drawings with 2000 strokes and four colors.}\vspace{-1.0em}
\label{fig:result_rob}
\end{figure}



Our new robotic drawing system relies on drawing strokes as underlying drawing primitives, namely {\em stroke-based rendering} or {\em stroke-based drawing} \cite{painterly}, as demonstrated in Fig.~\ref{fig:result_rob}. The primary goal of our robotic drawing is to make robotic stroke-based drawing artistic. At the same time, we would like to have the sequence of drawing strokes also look natural to human observers, which makes our system different from what can be obtained from a mechanical color printer. This aspect would be crucial for robotic art, particularly from the artistic performance point of view.

Our stroke-based rendering (SBR) method is guided by the semantic content in the image, but central to our idea is the use of depth information based on the following observation: Given a plausible depth map for a semantic unit in the image (\eg a face in a portrait) we distribute the strokes over the contours of the layers defined by a plane moving from the back of the face toward the viewer. As the drawing evolves, the viewer quickly perceives the intended shape and also gradually observes detail getting filled in. To avoid the appearance of uniformly shrinking curves, we use binning for the depth values, which appears to be quite successful in giving the impression that the focus is in different areas as the shape is being defined.

In summary, this paper makes the following contributions:

\begin{itemize}
\item New SBR system that integrates algorithms for image segmentation and depth estimation. The system does not require user assistance.

\item Algorithm for generating a stroke plan, \ie stroke ordering and placement, for a robotic painter. The aim is to convey to an observer of the drawing process a human-like painting style. The {\em depth map} is a central component of the proposed method, which we have named {\em layered depth}. 

\item We compare our approach against the state-of-the-art SBR methods and show that our results have a similar rendering quality but have a more natural stroke order.

\item We show various SBR-based artistic drawing results and demonstrate the robotic drawing of complicated artistic images using SBR and bi-manual coordinate parallel drawing.
\end{itemize}

The rest of this paper is organized as follows. We survey works relevant to SBR and robotic drawing in Sec.\ref{sec:related}.  In Sec.~\ref{sec:system}, we explain our SBR pipeline including stroke rendering and planning. Sec.~\ref{sec:robotdraw} depicts our stroked-based robotic drawing system.  We show our drawing results and provide a comparison with a state-of-the-art SBR method in Sec.~\ref{sec:results}, and conclude the paper in Sec.\ref{sec:conclusion}.

\section{PREVIOUS WORK}
\label{sec:related}

\subsection{Stroke-based Rendering}
Early SBR research was primarily concerned with the artistic quality of the final image. Litwinowicz~\cite{litwi} proposed a method for manipulating a reference image to achieve an impressionist effect. Strokes are generated starting from seed points placed on a regular grid by following the tangent to the gradient in both directions until a predefined length or an edge boundary. During rendering, the order of the strokes is randomized to avoid artifacts due to spatial coherence. Hertzmann's work~\cite{painterly} adapts this strategy by searching the neighborhoods of the stroke seeds for the point with the most significant error between the reference image and a smoothed version. In addition,~\cite{painterly}  considerably increases the range of possible styles by including a number of {\em style parameters} in the stroke generation and rendering process. A survey of early SBR work is given in~\cite{sbr-survey}.

While early research in SBR mainly considered local image properties, later work incorporated the semantic information by introducing user-guided or automatic image segmentation and labeling techniques. Lin et al.~\cite{painterly-video-lin} create a dictionary of brushes for different categories of objects and use the semantic content as a guide for stroke placement and rendering. Their system employs user-driven image segmentation. Hertzmann~\cite{painterly-video-hertzmann} presents a similar system that adds a higher degree of automation for the various tasks and integrates capabilities for the synthesis and editing of strokes. Zeng et al.~\cite{image-parsing} explore the semantic information via hierarchical decomposition by organizing the components in a parse tree that guides the rendering process. The segmentation subdivision is fairly detailed and includes both user-driven and automatic stages.

Advances in Machine Learning (ML) have generated renewed interest in SBR research by offering new ways to generate content. The systems can now "learn how to paint" via rules encoded in loss/reward functions but appear disconnected from the rendering process, unlike the early SBR research. Huang et al.~\cite{paint-DRL} use deep reinforcement learning (DRL) with adversarial loss feedback to learn a rendering sequence that minimizes the difference between the target image and the final painting. Singh and Zheng~\cite{semantic-DRL,intelli-paint} use a similar approach but also incorporate semantic information to create different patterns in processing foreground and background strokes. Their images are sharper, with a noticeable emphasis on the more refined features of the foreground objects.

The work in this paper shares the goal and motivation of Schaldenbrand and Oh~\cite{content-loss}. Their primary emphasis is on training the system to produce a stroke sequence that exhibits characteristics of human planning and can be executed by a robot arm. In other words, it should be evident early on what the system is painting, similar to the "blocking in" technique used by artists to quickly apply the rough outlines of the dominant elements in the scene. The kinematic constraint set is what sets apart our work and~\cite{content-loss} from the work in~\cite{paint-DRL,semantic-DRL}, which produces unrealistic strokes that a robotic system cannot execute.
While we share the same aim as~\cite{content-loss}, our approach is closer to that in~\cite{painterly-video-lin,painterly-video-hertzmann,image-parsing,semantic-DRL}. Our method is also guided by the semantic content in the image, but we use depth information to distribute the strokes over the contours to be gradually perceivable as the drawing evolves.


\subsection{Robotic Drawing}

Harold \etal did seminal artistic work on building drawing machines called AARON \cite{aaron90}. AARON can generate artwork using a plotting machine programmed by a computer.
Human portrait style drawing system was proposed using the HOAP2 humanoid~\cite{ceb05}.
Paul the robot \cite{tresset13} can draw a stylistic portrait by observing the target object with the camera and mimicking artistic signatures.
eDavid \cite{lpd13} uses a non-photorealistic algorithm in computer graphics for drawing and relies on an industrial robot with visual feedback.
Song \etal presented a robotic pen-art system called SSK that can create pen art on arbitrary surfaces using impedance control \cite{songICRA18} while minimizing drawing distortion \cite{songICRA19}. Liu \etal presented a robotic drawing system that can draw on a non-flat surface using closed-loop planning \cite{liu2021robust}. 
Generative art using machine learning is being increasingly used for robotic art. Wang \etal  proposed a drawing method for personalized avatar characters using generative adversarial network-based style transfer \cite{wang2020robocodraw}. Reinforcement learning is also explored to learn from the painter database where to place the brush strokes  \cite{content-loss}.
The main departure of our work from the existing robotic drawing systems is that we use stroke-based drawing and its ordering.

\section{STROKE-BASED RENDERING AND PLANNING}
\label{sec:system}

Given an input image $I$ and desired number of strokes $N$, our system generates a stroke plan for a robotic painter. Since a robotic painter may have a limited selection of colors (\eg pen-based system), the user can specify a target number of colors for the painting. In this case, we adopt the approach in~\cite{content-loss} that uses  $k$-means clustering to partition the RGB color space of the image into a palette of $k$ colors.

The proposed system integrates algorithms for instance and semantic segmentation (panoptic segmentation)~\cite{panoptic} and monocular depth estimation~\cite{midas}. State-of-the-art segmentation and depth analysis algorithms provide rich and reliable estimates about the image content that guide our selection of {\em stroke order} and {\em stroke placement}.
The following sections discuss each stage in the pipeline.

\subsection{Panoptic Segmentation}
\label{sec:panop}

\begin{figure*}[ht]
\centering
\subfigure[]{\includegraphics[height=68pt]{figs/targets/9.jpg}} 
\subfigure[]{\includegraphics[height=68pt]{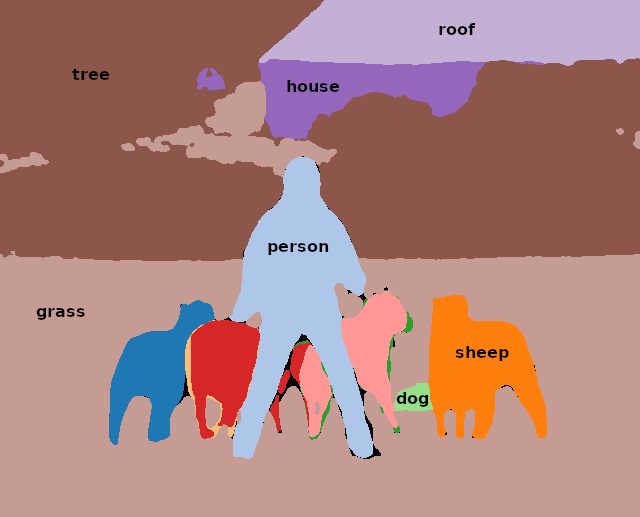}}
\subfigure[]{\includegraphics[height=68pt]{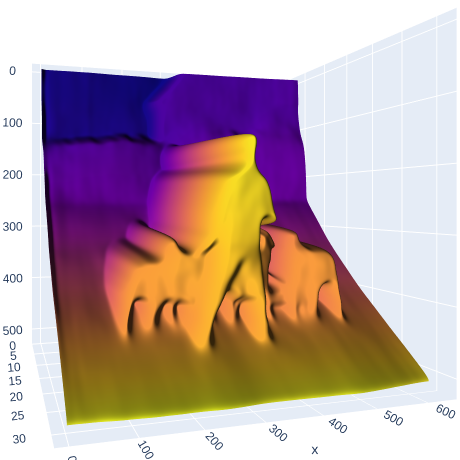}}
\subfigure[]{\includegraphics[height=68pt]{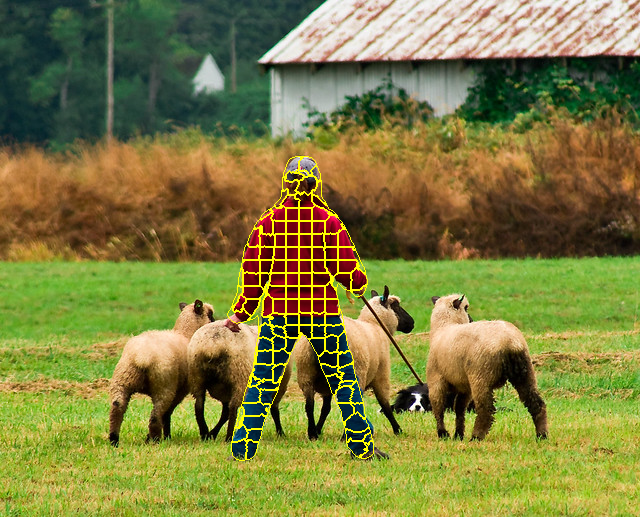}}
\subfigure[]{\includegraphics[height=68pt]{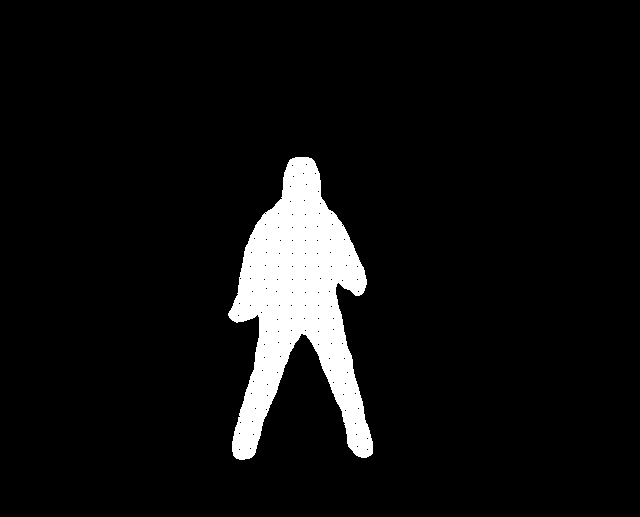}}
\subfigure[]{\includegraphics[height=68pt]{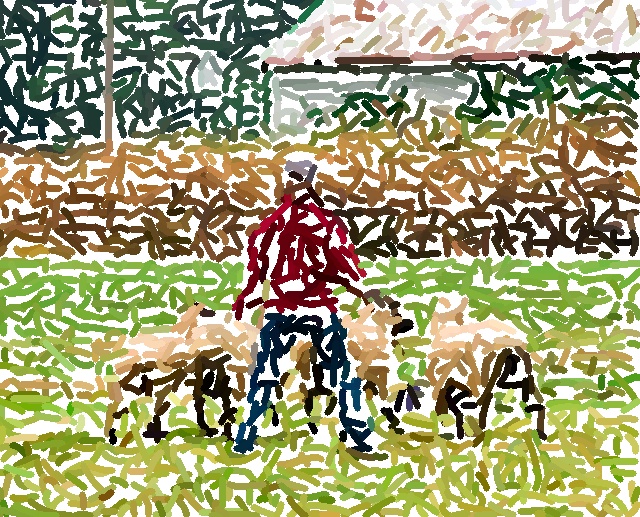}}
\caption{\label{fig:stages} The output from each stage in the pipeline: $(a)$ original image 641$\times$513; $(b)$ panoptic segmentation into things (person, sheep, dog) and stuff (house, roof, tree, grass); $(c)$ smoothed depth map; $(d)$ superpixel segmentation of the {\em person} prediction; $(e)$ stroke seeds for the {\em person} prediction based on superpixel segmentation; $(f)$ final painting with 2000 strokes of stroke-width six and no restrictions on number of colors.}
\end{figure*}

We use Detectron2~\cite{detectron2} to perform {\em panoptic segmentation}~\cite{panoptic} and assign semantic labels to the pixels in the image and to the identify the salient objects. The output of this stage is a union of two classes denoted in~\cite{panoptic} as {\em things} and {\em stuff} where the former represents the concrete objects in the scene. At the same time, the latter refers to areas that are less well defined but share similar textures or materials (\eg grass or snow). In the rest of the paper, we will use the term {\em prediction} to refer to both things and stuff. Fig.~\ref{fig:stages}$(b)$ shows a visual representation of the output from this stage on the original image in Fig.~\ref{fig:stages}$(a)$.

In our pipeline, things are processed first since they represent the concrete predictions, \ie salient elements, in the scene. In order to balance prediction confidence and instance dominance, each thing is assigned a weight that is the product of the reported confidence score and its area in pixels, \ie $thing.weight = thing.score \times thing.area$ and things are sorted by decreasing weight.

The stuff is processed in semantic order. We have reordered and grouped the semantic labels recognized by Detectron2 in the painting planning task.
The exact grouping and ordering within a group are subject to further investigation.

\subsection{Stroke Selection and Stroke Generation}
\label{sec:seeds}

For each prediction $p_i$ in the panoptic segmentation, we select a set of candidate pixels that serve as generators for the final strokes. Note that our main goal is to identify a good stroke ordering that could resemble the planning of a human painter. Thus, any established strategy for stroke selection and generation discussed in Sec.~\ref{sec:related} can be used at this stage.

In this work, the stroke seeds are chosen as the centroids of regions within $p_i$ that are perceptually similar. The computation of these regions, called {\em superpixels}, is essentially an image segmentation task considering the tradeoff between shape compactness (regularity) and perceptual similarity. The Simple Linear Iterative Clustering (SLIC) algorithm~\cite{slic1, slic2} is one of the prominent examples based on $k$-means that optimizes a distance function in 5D space which is a combination of the color distance in CIELAB and the spatial distance in the image. The particular implementation used in our work is based on the Watershed algorithm~\cite{watershed1,watershed2} from the Python {\em scikit-image} library~\cite{scikit}. Figs.~\ref{fig:stages}$(d)$~and~\ref{fig:stages}$(e)$, respectively, show the superpixel segmentation and the centroids of the corresponding regions for the {\em person} prediction.

Let $S_i$ represent the set of candidate seeds for prediction $p_i$. The size of $S_i$, \ie the number of superpixel regions to identify in $p_i$, is computed as $|S_i| = \frac{size(p_i)}{size(I)}N$, where $size(\cdot)$ returns the number of pixels within a region of the image. After identifying the set of stroke seeds, we use {\em AniPainter}~\cite{anipaint} to generate the strokes. AniPainter is a variation on the stroke generation algorithm proposed by Hertzmann~\cite{painterly}.

\subsection{Depth Estimation and Stroke Ordering}
\label{sec:depth}

We use MiDaS~\cite{midas} to perform depth estimation for the target image and then smooth the computed depth values with a Gaussian filter. Fig.~\ref{fig:stages}$(c)$ shows a visual representation of the smoothed depth map.

Next, we compute the depth map's histogram, $H$, and use the histogram's bins to determine the order of strokes. Algorithm~\ref{alg:seeds} outlines the process for a given prediction $p_i$. The algorithm partitions the set of stroke seeds $S_i$ into an ordered list of disjoint subsets $S_i'$, which we call {\em frames}. Referring to the analogy in Sec.~\ref{sec:intro} of a plane moving through the depth range for $p_i$ starting at the back and toward the viewer, each frame represents the stroke seeds in $S_i$ that lie in the intersection of the plane with $p_i$. Early in this process, the frames can be recognized as the silhouette of $p_i$. However, later on, there is enough variation within the depth map to give the impression of shifting attention to different areas in $p_i$. The histogram bins give "thickness" to the plane and ensure that for each frame, a sufficient number of candidate seeds are collected for painting.

Each frame is sorted using a two-level criterion. The image is subdivided into a 5x5 grid, and the candidate seeds are first ordered by their grid cells (top-to-bottom, left-to-right) and then within a cell by their $(i,j)$ coordinates. This aim is to direct the attention of the robot painter to a particular area for some time to avoid the appearance of scan-line processing. The coarse grid ensures that the robot painter will spend some time working on the right side of the "silhouette" of $p_i$ before switching to the left side.

\begin{algorithm}
\SetKwInOut{Input}{Input} 
\Input{$H$, depth map histogram for target image $I$ \\
$S_i$, set of stroke seeds for prediction $p_i$
}
\KwOut{set of frames to be painted for $p_i$}

$frames \gets [~]$

\ForEach{$bin \in H$}{
  $S'_i \gets bin \cap S_i$\\
  sort $S_i'$ by pixel coordinates and 5$\times$5 grid\\
  $frames.append(S_i')$
}
\Return{frames}
\caption{Stroke Ordering}
\label{alg:seeds}
\end{algorithm}

\subsection{Layered Depth Algorithm}

The algorithm for generating a stroke plan to produce a painting for image $I$ given a number of strokes $N$ is shown in Algorithm~\ref{alg:main}. We begin by computing the panoptic segmentation to identify the salient objects ({\em things}) and regions of interest ({\em stuff}). Objects in the thing class are sorted by {\em weight} that considers both the size of the item and the confidence score of the prediction. We use the semantic ordering 
for objects in the stuff class. Next, we compute the depth map and its histogram, which are the critical components of the proposed method. They enable us to generate a stroke ordering that conveys deliberate planning to an outside observer. For each object in the panoptic segmentation, we compute a sequence of frames of candidate stroke seeds. The back-to-front order ensures that the strokes are reasonably far apart in the beginning, giving the impression that the robot artist is sketching the object outline. As the frames move to the front, variations in the depth map cluster the strokes into separate sub-regions, giving the appearance that the robot is concentrating on the features of the object.

\begin{algorithm}
\SetKwInOut{Input}{Input} 
\Input{$I$, target image $I$\\
$N$, total number of strokes
}
\KwOut{$S$, set of stroke seeds for painting of $I$}

$P \gets PanopticSegmentation(I)$  \hfill ~\ref{sec:panop}\\
sort $P$ by $w_i=score_i \times area_i$  (for {\em things}) and semantic meaning (for {\em stuff}) \\

$D \gets ComputeDepthMap(I)$ \hfill ~\ref{sec:depth}\\
$H \gets ComputeHistogram(D)$\\

$frames \gets [~]$\\
\ForEach{$p_i \in P$}{
  $S_i \gets ComputeStrokeSeeds(p_i)$ \hfill ~\ref{sec:seeds}\\
  $frames_i \gets ComputeFrames(H, S_i)$ \hfill Alg.~\ref{alg:seeds}\\
  $frames.append(frames_i)$\\
}
\Return{frames}
\caption{Layered Depth Painting}
\label{alg:main}
\end{algorithm}

\begin{figure*}[ht!]
\centering
\begin{tabular}{cccccc}
Target & 50 strokes & 250 strokes & 500 strokes & 1000 strokes & Painting \\


\frame{\includegraphics[width=60pt]{figs/targets/6.jpg}} &
\frame{\includegraphics[width=60pt]{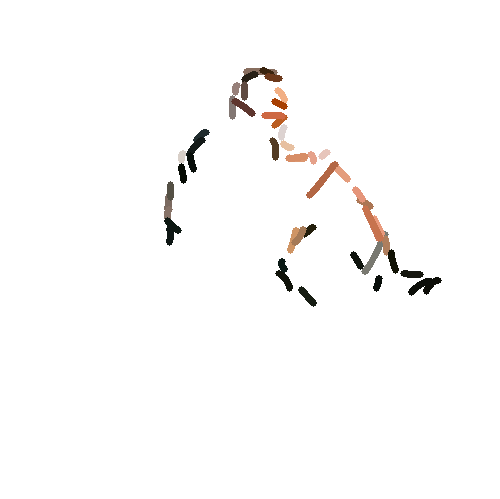}} &
\frame{\includegraphics[width=60pt]{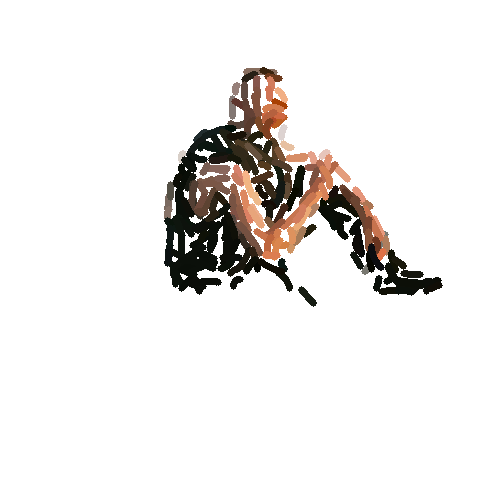}} &
\frame{\includegraphics[width=60pt]{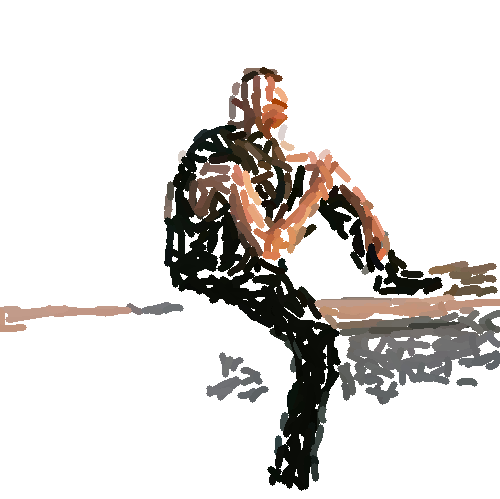}} &
\frame{\includegraphics[width=60pt]{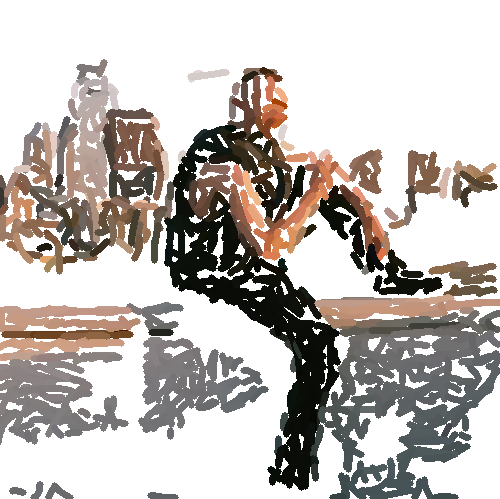}} &
\frame{\includegraphics[width=60pt]{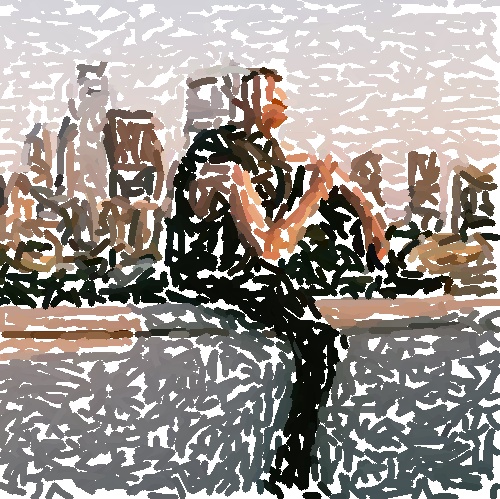}} \\

\frame{\includegraphics[width=60pt]{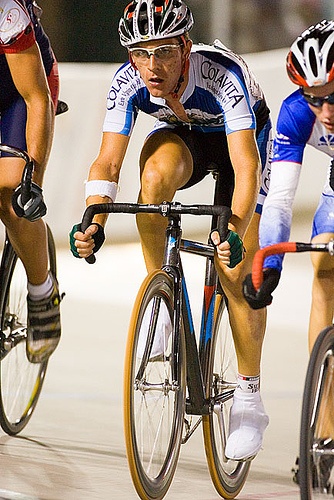}} &
\frame{\includegraphics[width=60pt]{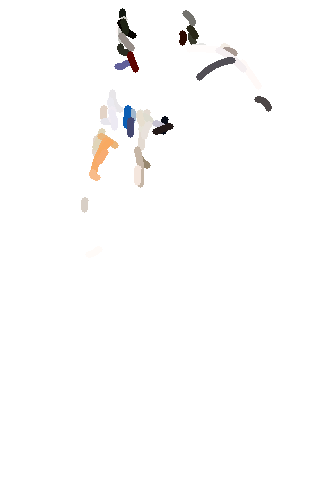}} &
\frame{\includegraphics[width=60pt]{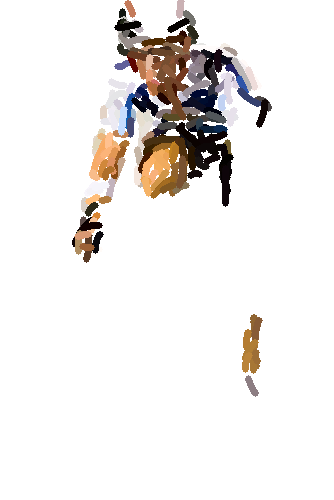}} &
\frame{\includegraphics[width=60pt]{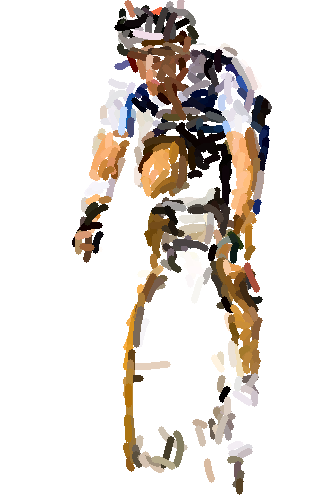}} &
\frame{\includegraphics[width=60pt]{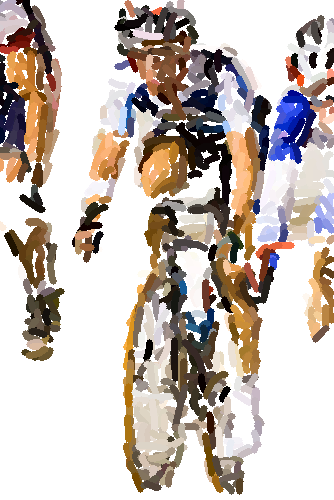}} &
\frame{\includegraphics[width=60pt]{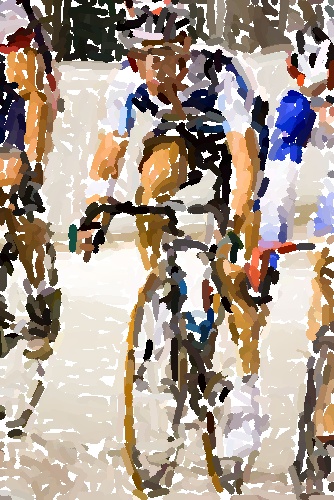}} \\

\frame{\includegraphics[width=60pt]{figs/targets/9.jpg}} &
\frame{\includegraphics[width=60pt]{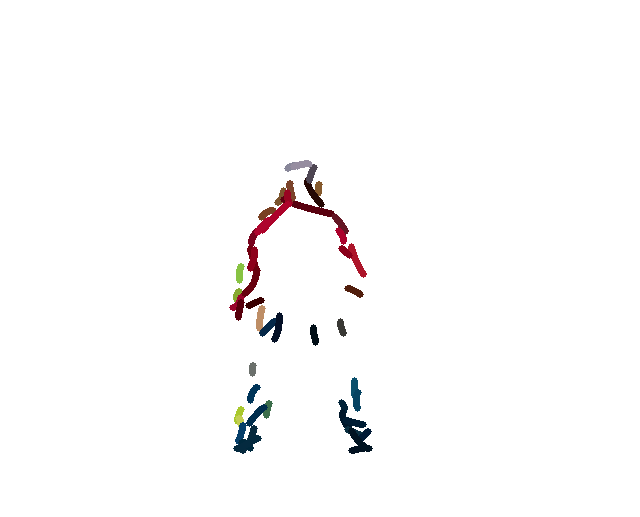}} &
\frame{\includegraphics[width=60pt]{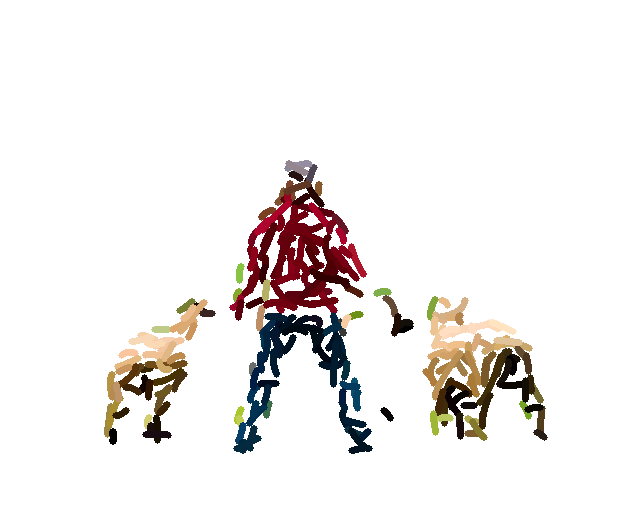}} &
\frame{\includegraphics[width=60pt]{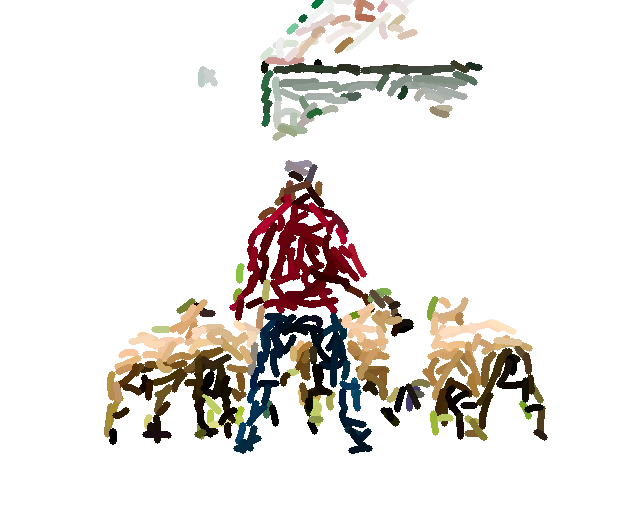}} &
\frame{\includegraphics[width=60pt]{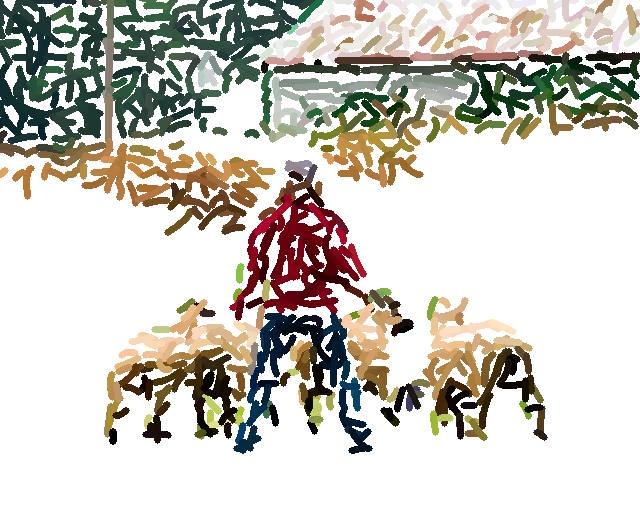}} &
\frame{\includegraphics[width=60pt]{figs/colors-all/9/painting.jpg}} \\

\frame{\includegraphics[width=60pt]{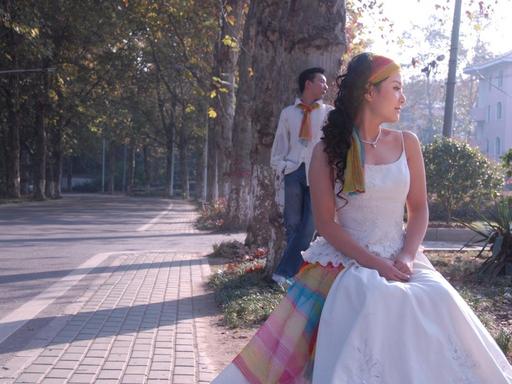}} &
\frame{\includegraphics[width=60pt]{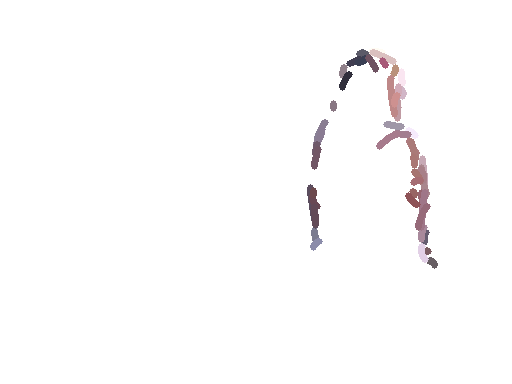}} &
\frame{\includegraphics[width=60pt]{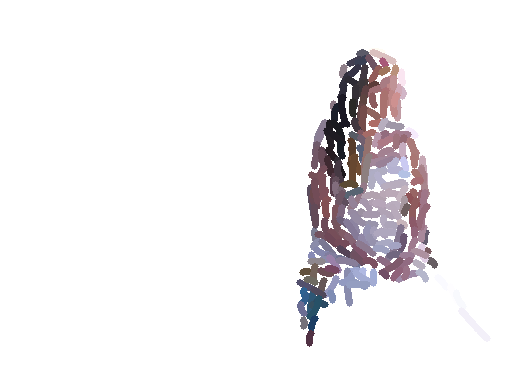}} &
\frame{\includegraphics[width=60pt]{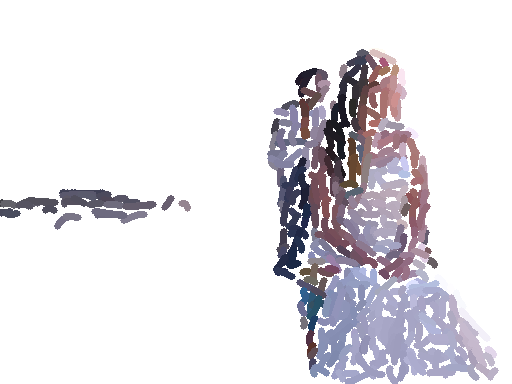}} &
\frame{\includegraphics[width=60pt]{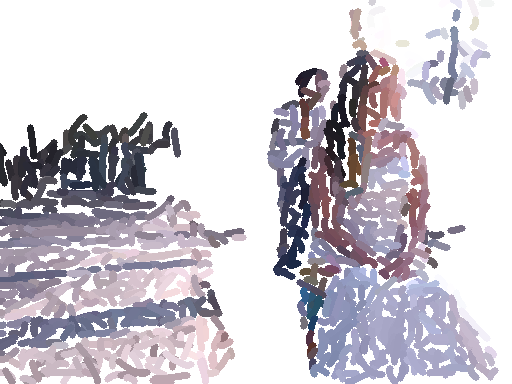}} &
\frame{\includegraphics[width=60pt]{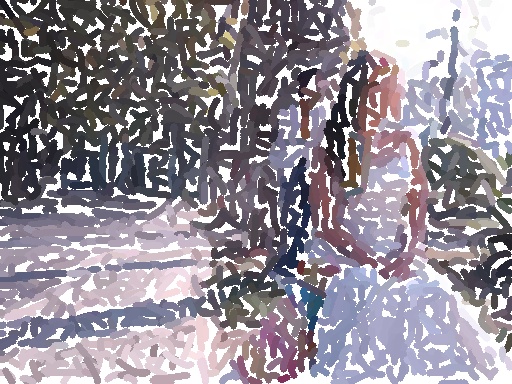}} \\

\frame{\includegraphics[width=60pt]{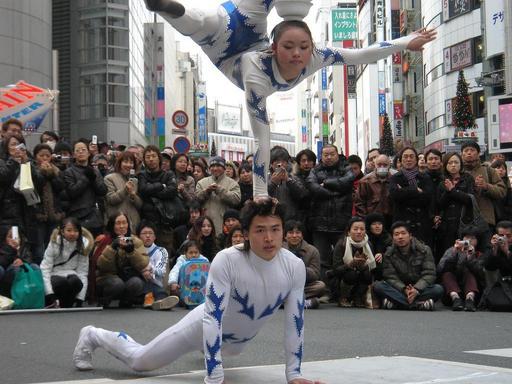}} &
\frame{\includegraphics[width=60pt]{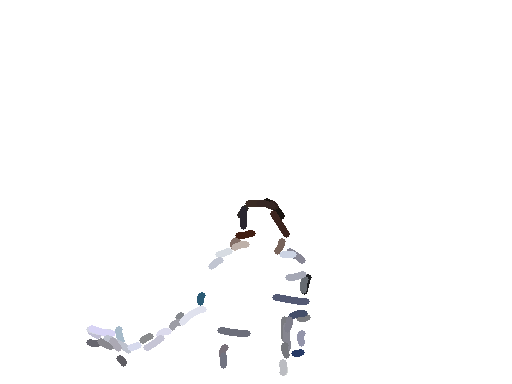}} &
\frame{\includegraphics[width=60pt]{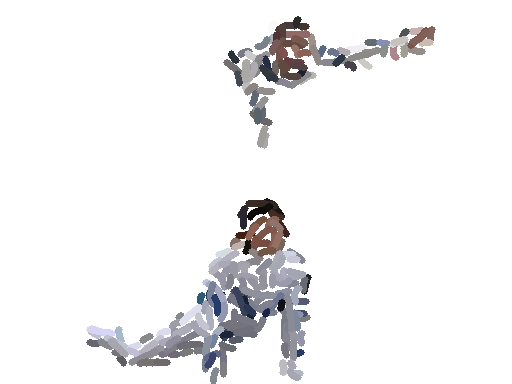}} &
\frame{\includegraphics[width=60pt]{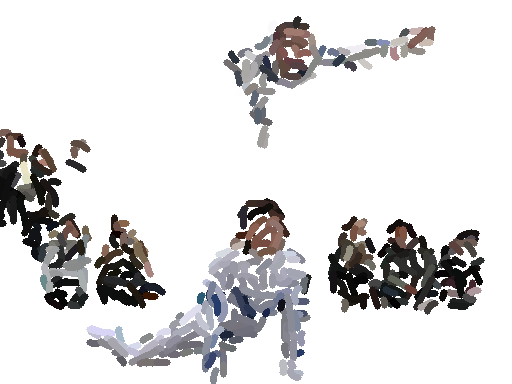}} &
\frame{\includegraphics[width=60pt]{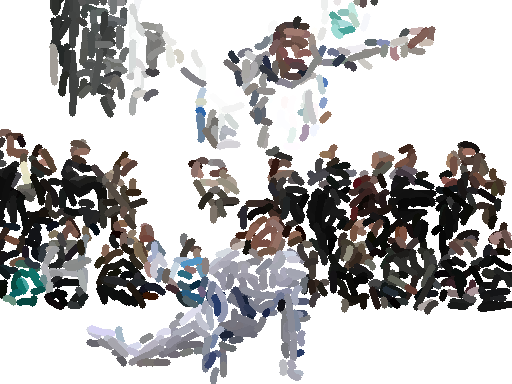}} &
\frame{\includegraphics[width=60pt]{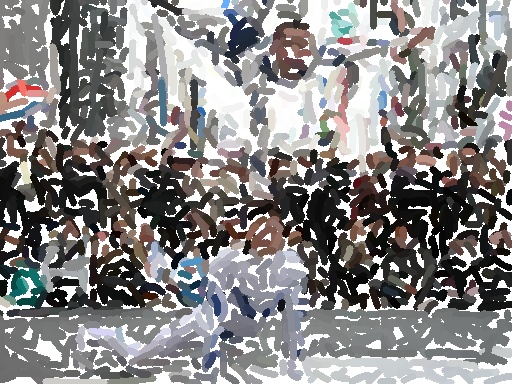}} \\
\end{tabular}
\caption{\label{fig:results} Representative results for $N=2000$. All strokes are opaque of fixed stroke-width 6 with no restriction on the number of colors. The stroke sequence is determined by Algorithm~\ref{alg:main} and the strokes are generated and rendered by AniPainter~\cite{anipaint}.}
\end{figure*}

\section{Stroke-based Robotic Drawing}
\label{sec:robotdraw}

\begin{figure}[htb]
\centering
\includegraphics[height=1.6in]{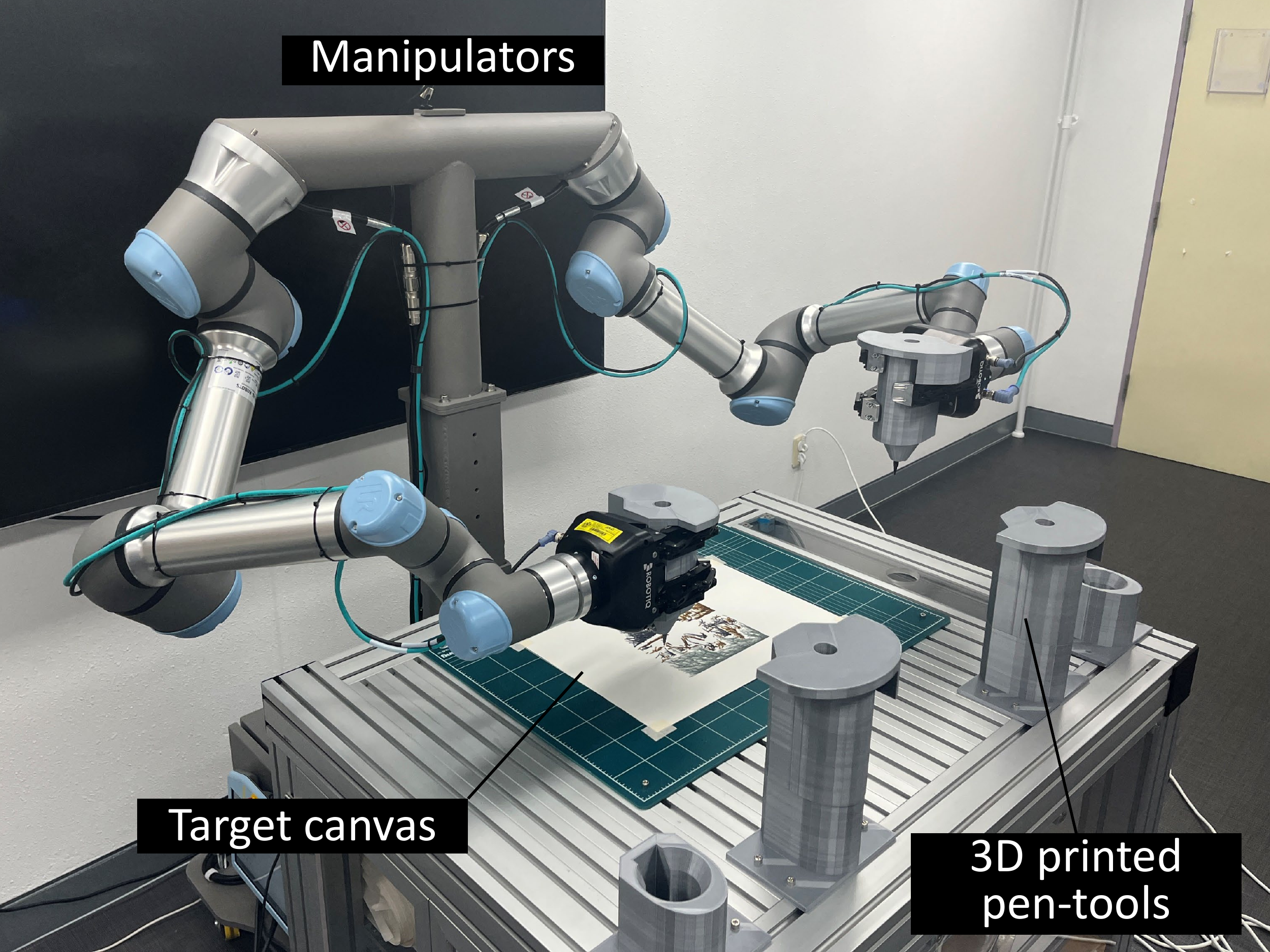}\hspace{0.2em}\includegraphics[height=1.6in]{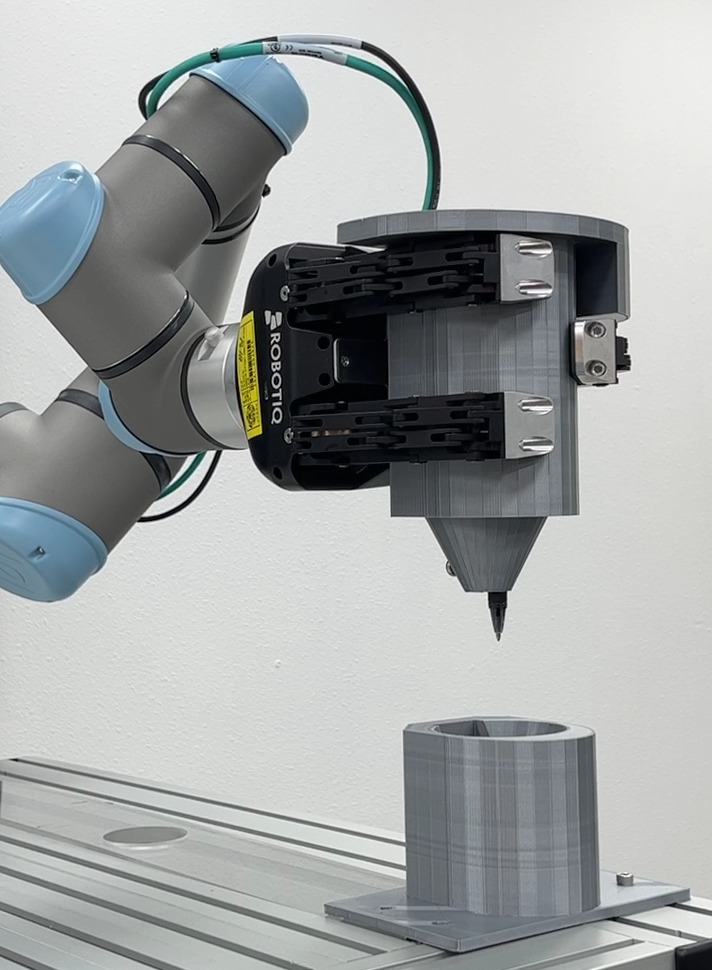}
\caption{Bi-manual robotic drawing system setup with pen-tool change mechanism.} 
\label{fig:setup}
\end{figure}

We render our simulated strokes in a physical space using a robot painter. As shown in Fig.~\ref{fig:setup}, our robotic setup consists of two UR5e manipulators, each equipped with Robotiq adaptive 3-finger gripper to hold the pen tool. The gripper allows us to change the colors whenever needed, resulting in a fully automated robotic drawing system. Our novel design of a pen-holding tool enables our system to robustly change the tools even under slight motion perturbations due to position or motion errors. 

\subsection{Bi-manual Coordinated Parallel Drawing}

The robotic manipulator drawing the pen strokes can be viewed as a problem finding continuous robot joint configurations whose end-effector follows the given stroking path. We solve this problem by iteratively solving the Inverse Kinematics (IK) for the given drawing path~\cite{chitta2012moveit,beeson2015trac}. 
We set the minimum distance objective in the configuration space to ensure no sudden joint jumps happen, which can result in bad drawing results or collision with the drawing canvas. 

Although many possible drawing strategies exist for dual-arm setup for efficient robotic drawing, we aim to draw using dual manipulators sharing the same drawing canvas space $\mathcal{C}$. 
Because the canvas is shared, collision-aware robot motion planning is mandatory for drawing strokes, considering each manipulator moving in real-time. However, this is a costly and not scalable solution for our system, which draws thousands of strokes. Instead, we perform a {\em coordinated parallel drawing} between the manipulators. 
Specifically, we perform bi-manual drawing by sub-dividing the canvas into half, $\mathcal{C}_{right}$ and $\mathcal{C}_{left}$, where $\mathcal{C}_{right}$ represents the canvas space that is on the right manipulator side. We take the drawing sequence as the priority. 
{When the right manipulator draws on $\mathcal{C}_{right}$, the left manipulator will initiate the first stroke in the sequence on $\mathcal{C}_{left}$ and execute it concurrently with the right manipulator.}
We pre-plan the robot trajectory for each manipulator's tool change that does not interfere with the other one's drawing canvas space so that the tool change can also be performed whenever necessary.

\section{EXPERIMENTAL RESULTS}
\label{sec:results}

\subsection{Stroke Rendering Results}

Fig.~\ref{fig:results} shows representative results on images of varying complexity and composition. Guided by the segmentation algorithm, we can focus on the prominent objects in the scene, starting with a rough outline for each and progressively filling in details. The depth map contains enough variation that, combined with the grid-based sorting, gives the appearance of human-like planning of sketching different parts of the silhouette or concentrating on different areas within the object or within the scene when drawing background elements.

Table~\ref{tbl:stats} gives a summary of execution times for the various stages in the pipeline. The experiments were run on Ubuntu 22.04-based machine with NVIDIA GeForce RTX 2060 GPU (CUDA 11.7; Driver 515.48.07) and Intel\textsuperscript{\textregistered} Core\textsuperscript{\texttrademark} i9-9900 CPU @ 3.10GHz $\times$ 16 CPU.

\begin{table}[h]
\caption{Execution Times in Seconds for the Results in Fig.~\ref{fig:results}}
\label{tbl:stats}
\begin{center}
\begin{tabular}{|l|c|c|c|}
\hline
 Target ($W \times H$) & Segmentation & Depth & Overall \\
\hline
Row 1 ($500 \times 499$)~\cite{content-loss-repo} & 7.97 & 4.09 & 106.25 \\
Row 2 ($334 \times 500$)~\cite{voc12} & 7.99 & 4.18 & 85.05 \\
Row 3 ($641 \times 513$)~\cite{coco} & 17.30 & 4.33 & 148.95 \\
Row 4 ($512 \times 384$)~\cite{image-parsing} & 8.83 & 4.12 & 80.31 \\
Row 5 ($512 \times 384$)~\cite{mit300} & 18.91 & 4.26 & 90.61 \\
\hline
\end{tabular}
\end{center}
\end{table}

\begin{figure*}[ht!]
\centering
\begin{tabular}{cccccc}
Target & 50 strokes & 250 strokes & 500 strokes & 1000 strokes & Painting \\
\frame{\includegraphics[width=60pt]{figs/targets/1.jpg}} &
\frame{\includegraphics[width=60pt]{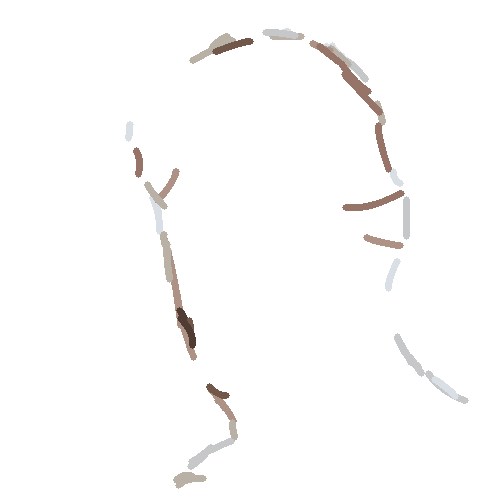}} &
\frame{\includegraphics[width=60pt]{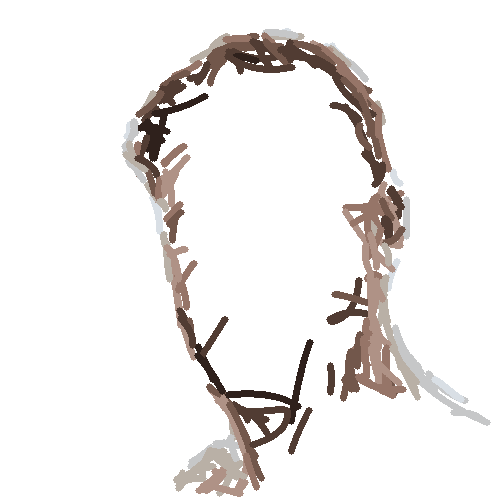}} &
\frame{\includegraphics[width=60pt]{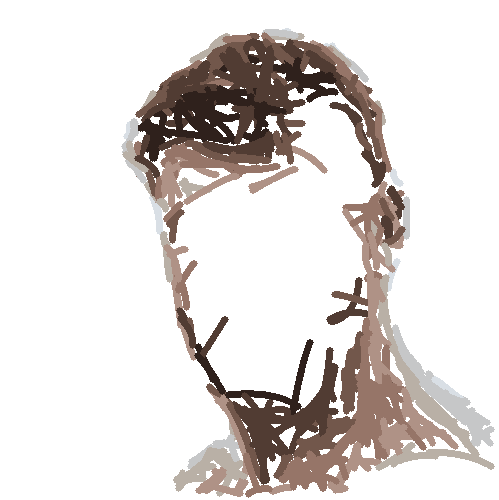}} &
\frame{\includegraphics[width=60pt]{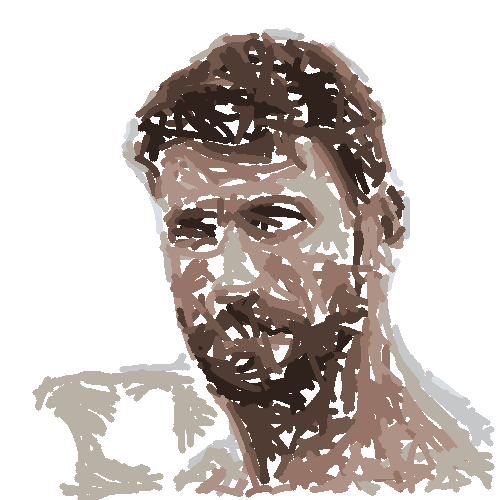}} &
\frame{\includegraphics[width=60pt]{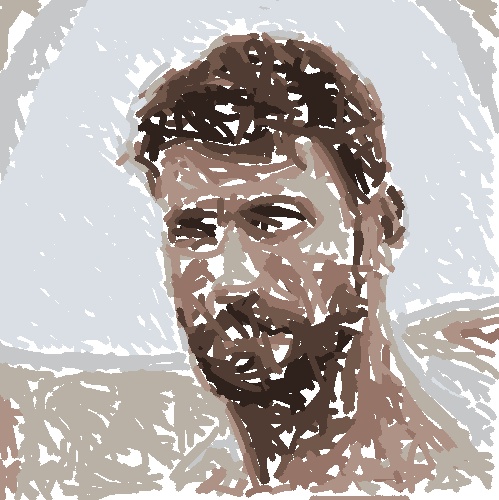}} \\
\vbox to 1pt {\vfil \hbox to 60pt{} \vfil } &
\frame{\includegraphics[width=60pt]{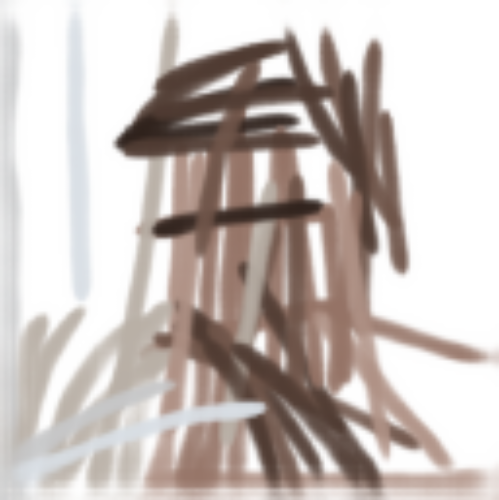}} &
\frame{\includegraphics[width=60pt]{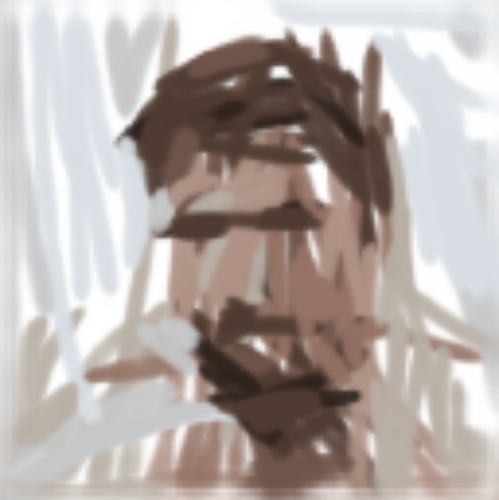}} &
\frame{\includegraphics[width=60pt]{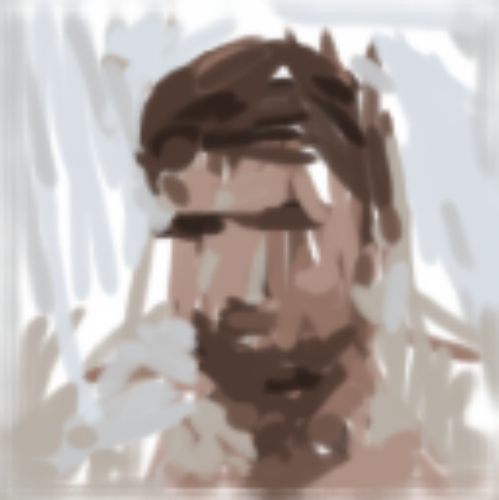}} &
\frame{\includegraphics[width=60pt]{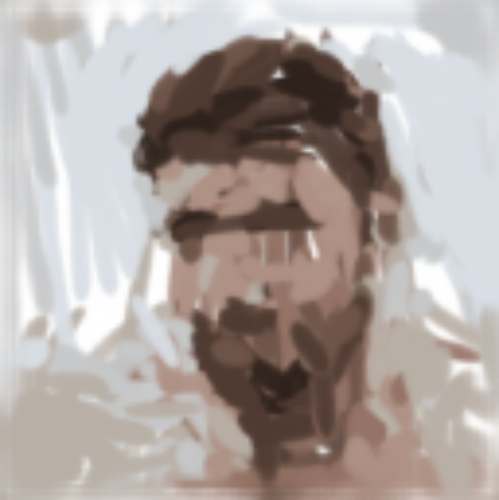}} &
\frame{\includegraphics[width=60pt]{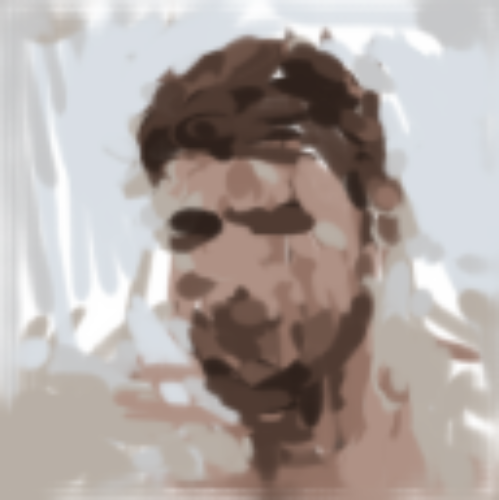}} \\
\frame{\includegraphics[width=60pt]{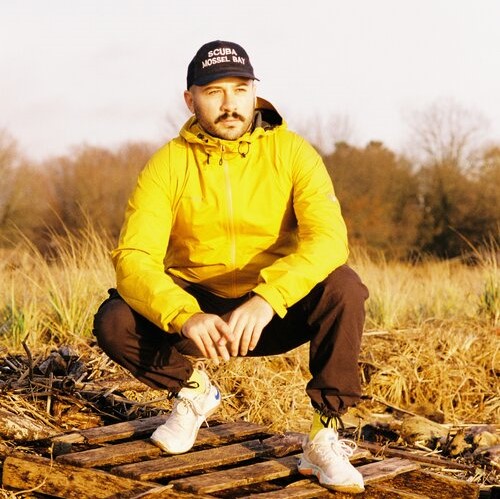}} &
\frame{\includegraphics[width=60pt]{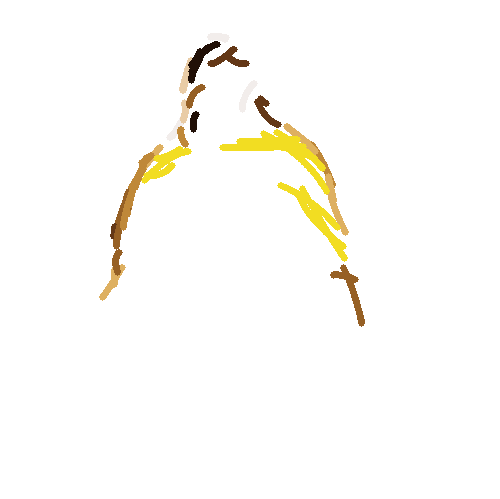}} &
\frame{\includegraphics[width=60pt]{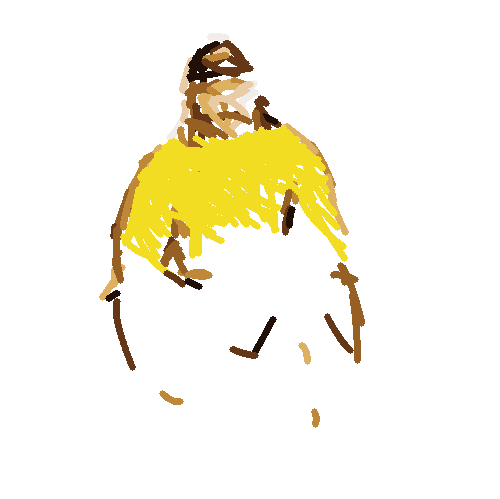}} &
\frame{\includegraphics[width=60pt]{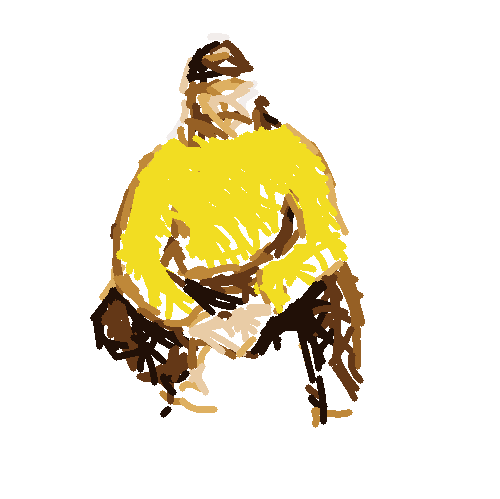}} &
\frame{\includegraphics[width=60pt]{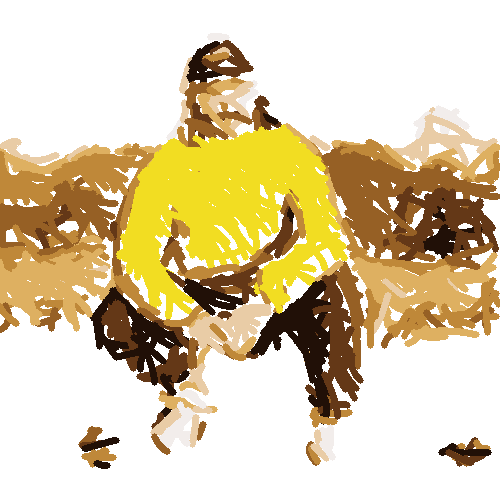}} &
\frame{\includegraphics[width=60pt]{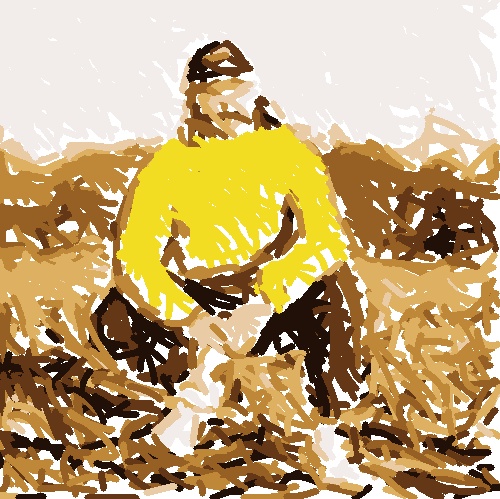}} \\
\vbox to 1pt {\vfil \hbox to 60pt{} \vfil } &
\frame{\includegraphics[width=60pt]{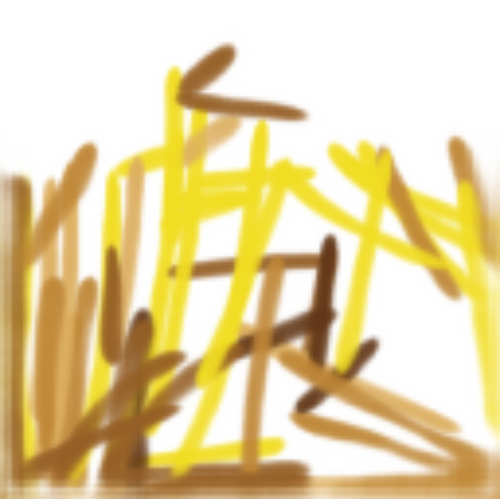}} &
\frame{\includegraphics[width=60pt]{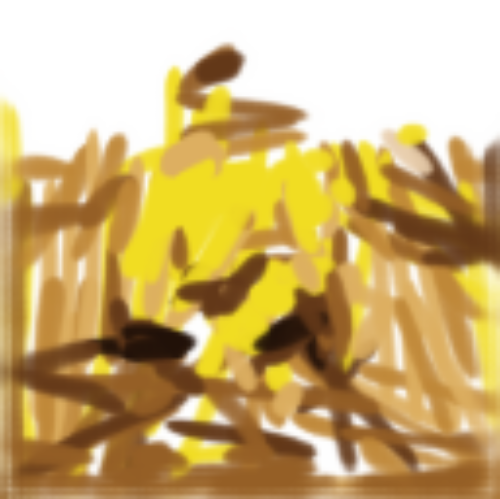}} &
\frame{\includegraphics[width=60pt]{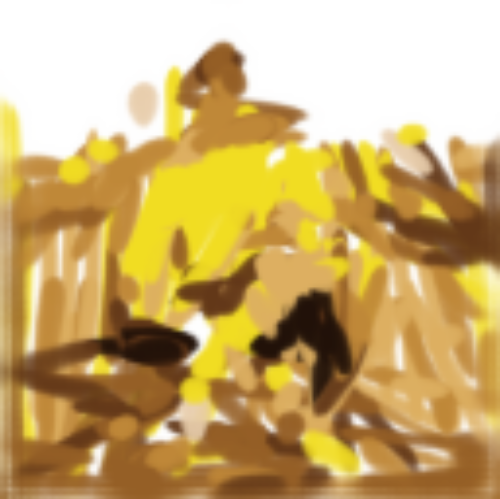}} &
\frame{\includegraphics[width=60pt]{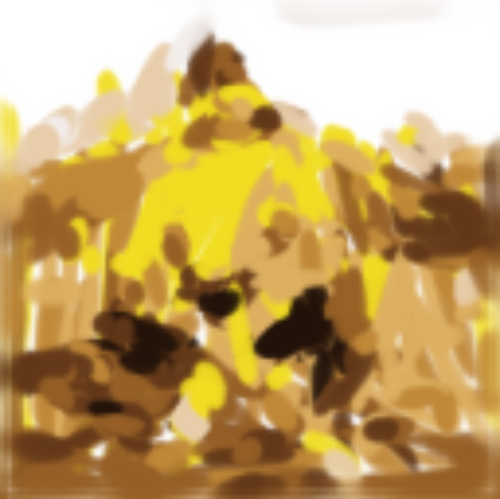}} &
\frame{\includegraphics[width=60pt]{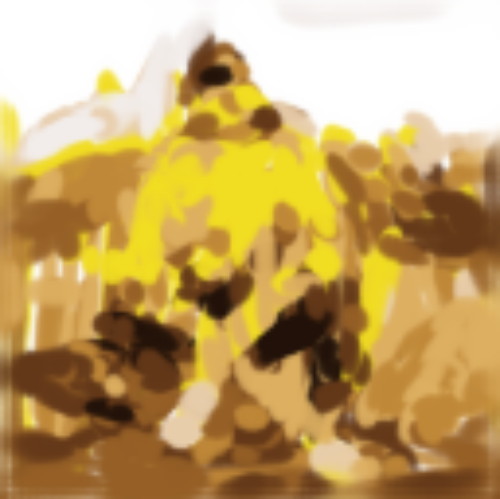}} \\
\frame{\includegraphics[width=60pt]{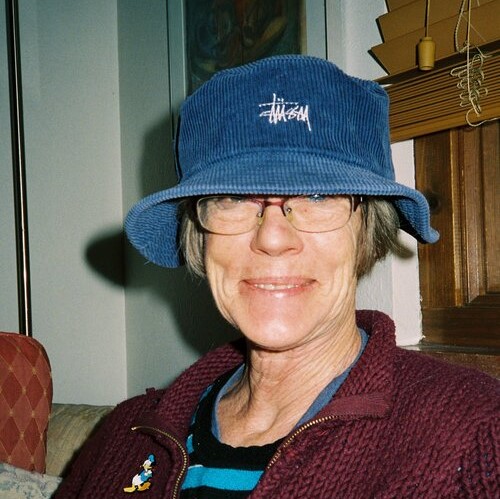}} &
\frame{\includegraphics[width=60pt]{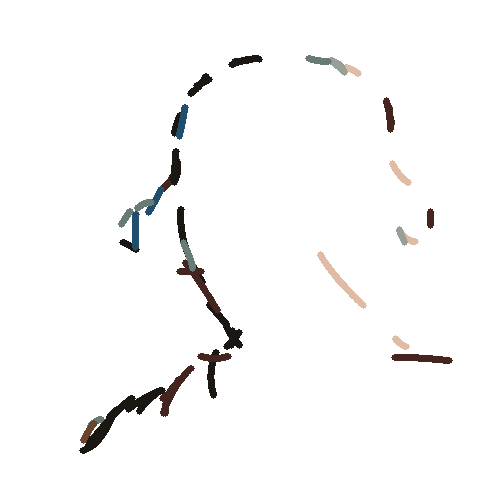}} &
\frame{\includegraphics[width=60pt]{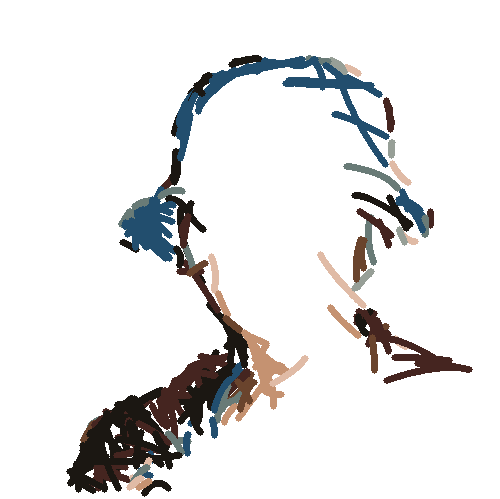}} &
\frame{\includegraphics[width=60pt]{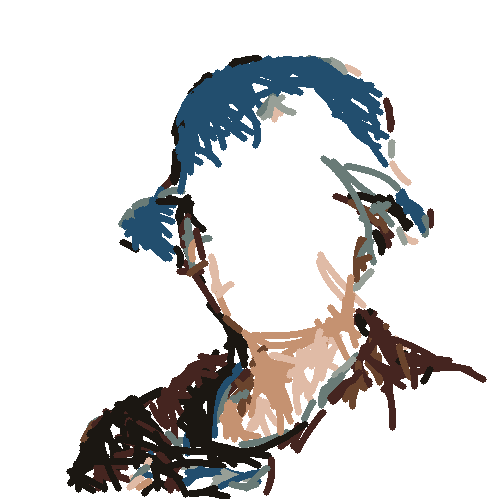}} &
\frame{\includegraphics[width=60pt]{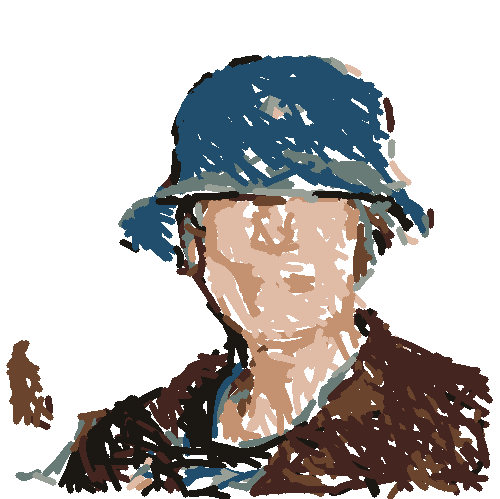}} &
\frame{\includegraphics[width=60pt]{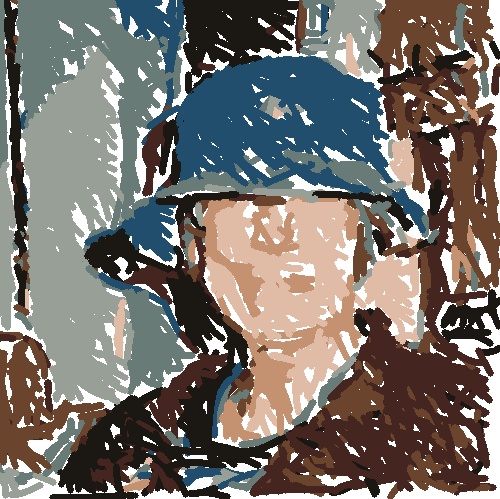}} \\
\vbox to 1pt {\vfil \hbox to 60pt{} \vfil } &
\frame{\includegraphics[width=60pt]{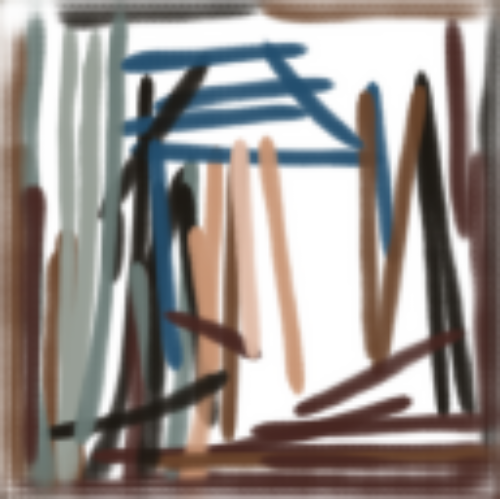}} &
\frame{\includegraphics[width=60pt]{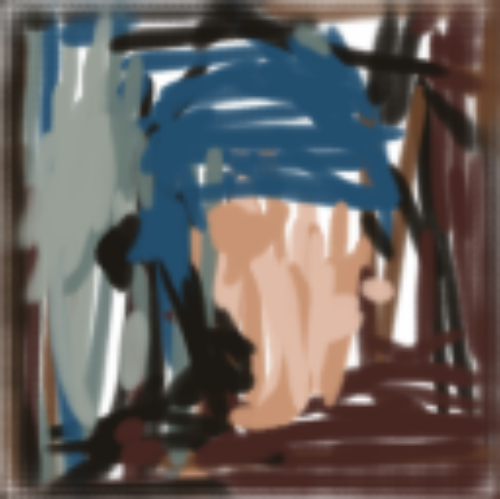}} &
\frame{\includegraphics[width=60pt]{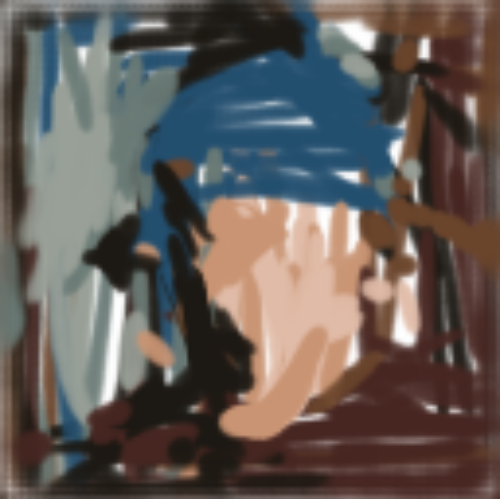}} &
\frame{\includegraphics[width=60pt]{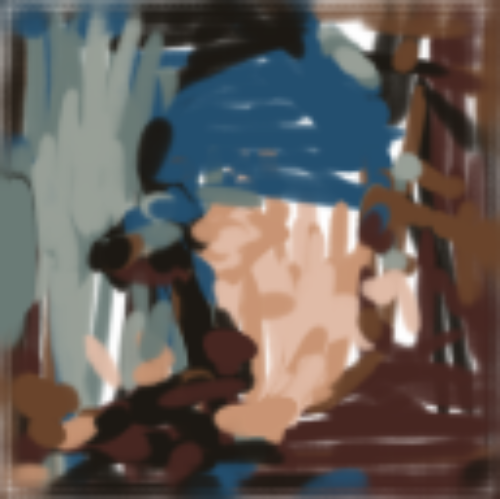}} &
\frame{\includegraphics[width=60pt]{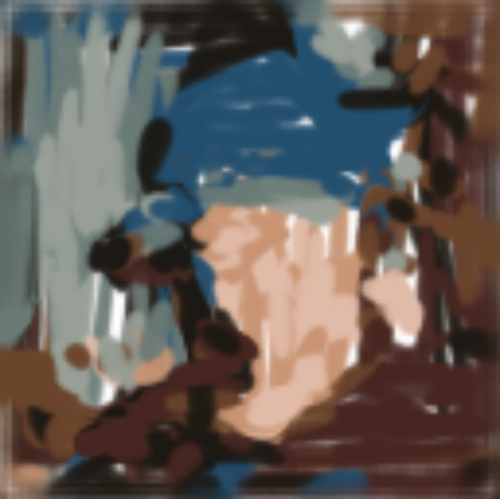}} \\
\end{tabular}
\caption{\label{fig:compare} Comparison between the results from our work (top row) with the results in~\cite{content-loss} (bottom row) for $N=2000$ and a palette of 8 colors. The target images are from the software repository~\cite{content-loss-repo} that accompanies~\cite{content-loss}. This is not so much a direct comparison, but rather a juxtaposition of alternative approaches. For example, the model in~\cite{content-loss} was trained for wider and longer strokes, which gives the effective "blocking in" appearance early on. We choose a narrower brush that helps sketch out the outline and fill in details, but requires more strokes to cover the canvas.}
\vspace*{-0.5em}
\end{figure*}

Fig.~\ref{fig:compare} shows a comparison with the results of the Content Masked Loss (CML) painter of Schaldenbrand and Oh~\cite{content-loss}. CML targets portrait-like scenes and successfully achieves the "blocking in" effect employed by an artist using large strokes to quickly compose the scene before filling in the details with strokes that tend to get smaller. Currently, our method has a less global reach in the early stages of the painting process but exhibits more deliberate behavior when working on the finer details. In some sense, a one-to-one comparison may not be relevant here since the two methods could be seen as having different painting/drawing styles. In both cases, the stroke sequence appears planned and purposeful.

\subsection{Robotic Rendering Results}

Fig.~\ref{fig:result_rob} shows our robotic rendering results; also refer to our digital rendering results shown in Figs.~\ref{fig:results} and~\ref{fig:compare}. We generated the data with 2000 strokes with four colors. The paper drawing canvas has a physical dimension of $160 \times 160 \sim 160 \times 200$ $\mathrm{mm}^2$. 
The overall robotic drawing by our system takes hours of time because we executed our robot at a low speed. This decision avoids any possible hazardous situations in the robot or the surrounding environment (\ie robot operator). 
As seen from the result, the colors of the digital and physical robotic rendering results are not the same. This is because the colors of the drawing tool (pen) do not match the color palette we obtained using our RGB color-clustering algorithm. 
Although we have used four colors for experiments, our system can handle more colors if we 3d-print more pen tools and use the full reachability of both manipulators for the tool change. 
The white spaces come from the stroke width difference for the pen. Instead, reducing the target drawing size or using a thicker pen would help reduce these spaces. The experiment revealed that 2000 strokes were too much for our current experimental setup considering the canvas and pen stroke sizes. The drawing stroke tends to be applied to a similar place multiple times, sometimes tearing the paper. Considering its thickness and color, finding a proper drawing tool is a current challenge in our system. 



\section{CONCLUSIONS}
\label{sec:conclusion}

In this paper, we present a new robotic drawing system based on stroke-based rendering that can create drawings resembling the input image artistically and the sequence of the strokes that human artists would create. Our SBR system is based on image segmentation and depth estimation to generate drawing strokes so that humans can perceive the intended shape quickly but gradually better when observed. 
We demonstrate that our SBR-based drawing makes artistic images, and our robotic system can replicate the result that human artists would draw.  {While the motion of our robotic drawing system may not resemble that of a human artist, the system's ability to emulate the drawing sequence of a human is of greater importance in this paper. Thus, our system has the potential to serve as both an automated drawing tool and as an artistic performance for human artists to experiment with. }

In future work, we plan to experiment with other ordering sequences at the global level. Currently, we process the elements in the scene by the {\em weight} property defined in Sec.~\ref{sec:panop}. An alternative approach might be to draw the first few layers for each element in the {\em things} class. Since the first few layers tend to show the element outline, this would give the appearance of pre-planning the scene composition, which might be particularly effective for a complex scene such as the one in the bottom row of Fig.~\ref{fig:results}. 



\addtolength{\textheight}{-4cm}   






\bibliographystyle{IEEEtran.bst}
\bibliography{layered_depth}

\end{document}